\documentclass{article}
\usepackage{ijcai22}

\usepackage{times}
\usepackage{soul}
\usepackage{url}
\usepackage[hidelinks]{hyperref}
\usepackage[utf8]{inputenc}
\usepackage[small]{caption}
\usepackage{graphicx}
\usepackage{amsmath}
\usepackage{amsthm}
\usepackage{booktabs}
\usepackage{multirow}
\usepackage{microtype}
\usepackage{wrapfig}
\usepackage{subcaption}
\usepackage{paralist} 
\usepackage[dvipsnames]{xcolor}
\usepackage{pifont}
\newcommand{\cmark}{\color{Green}\ding{51}}%
\newcommand{\xmark}{\color{red}\ding{55}}%

\newcommand{\task}[1]{\textsf{\small #1}}
\usepackage{fancyhdr}
\title{What is Right for Me is Not Yet Right for You:\\ A Dataset for Grounding Relative Directions via Multi-Task Learning}

\author{%
  Jae Hee Lee$^1$\footnote{Core contributors. JHL lead the research and contributed to evaluations and analyses. MK  contributed to the generation of the GRiD-3D scenes. KA contributed to the generation of the GRiD-3D questions. All three contributed to designing the research and writing the manuscript and all authors discussed and revised the manuscript. 
} \and Matthias Kerzel$^{1*}$\and Kyra Ahrens$^{1*}$\and Cornelius Weber$^1$\And Stefan Wermter$^1$%
  \affiliations%
  $^1$University of Hamburg%
  \emails%
  \{jae.hee.lee, matthias.kerzel, kyra.ahrens, cornelius.weber, stefan.wermter\}@uni-hamburg.de%
}

\begin{document}

\maketitle
\thispagestyle{fancy}
\begin{abstract}
  Understanding spatial relations is essential 
  for intelligent agents 
  to
  act and communicate in the physical world. Relative directions are spatial relations that describe the relative positions of target objects with regard to the intrinsic orientation of reference objects. 
  Grounding relative directions is more difficult than grounding absolute directions because it not only requires a model to detect objects in the image and to identify spatial relation based on this information, but it also needs to recognize the orientation of objects and integrate this information into the reasoning process.
  %
  %
  We investigate the challenging problem of grounding relative directions with end-to-end neural networks. To this end, we provide GRiD-3D, a novel dataset  that features relative directions and complements
  existing visual question answering (VQA) datasets, such as CLEVR, that involve only absolute directions. We also provide baselines for the dataset with 
  two
  established end-to-end VQA models. 
  Experimental evaluations
  show that answering questions on relative directions is feasible when questions in the dataset simulate the necessary subtasks for grounding relative directions. We 
  discover that those subtasks are learned in an order that reflects the steps of an intuitive pipeline for processing relative directions.
\end{abstract}

\section{Introduction}
\label{sec:introduction}

\begin{figure}[t]
  \centering%
  \hfill
  \begin{minipage}{0.6\linewidth}
    \begin{subfigure}[t]{\linewidth}
      \centering
      \includegraphics[width=\linewidth,trim={0cm 0cm 1cm 0cm},clip]{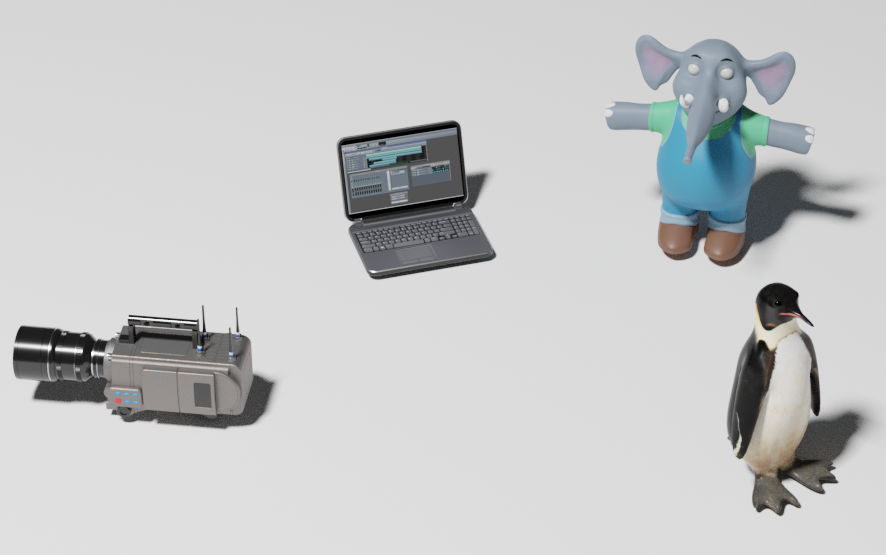}
      \caption{An example scene where an object \textit{laptop} is \textit{to the right} of another object \textit{elephant}. Here, the directional relation \textit{right} is relative and is determined by the perspective of the \textit{elephant}. \label{fig:horse_penguin}}
    \end{subfigure}
  \end{minipage}
  \hfill
  \begin{minipage}{0.389\linewidth}
    \begin{subfigure}[t]{\linewidth}
      \centering%
      \includegraphics[width=0.6\linewidth,trim={10cm 9cm 0cm 0.24cm},clip]{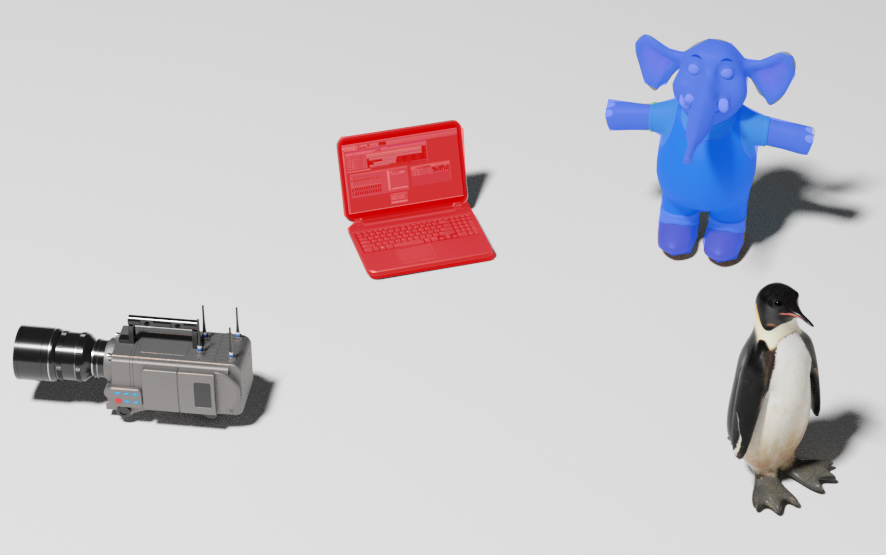}%
      \caption{Object detection\label{fig:subtask1}}
    \end{subfigure}\medskip\\
    \begin{subfigure}[t]{\linewidth}
      \centering%
      \includegraphics[width=0.6\linewidth,trim={10cm 7cm 0cm 0.24cm},clip]{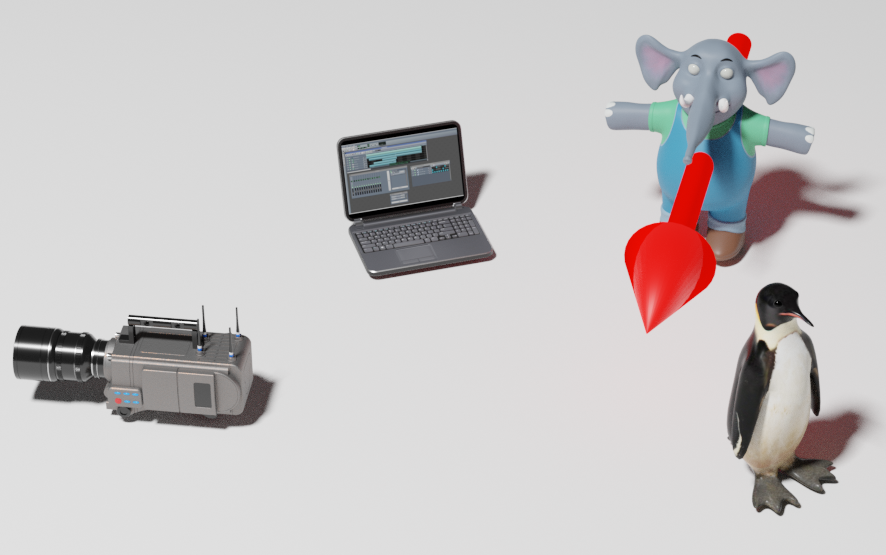}%
      \caption{Orientation estimation\label{fig:subtask2}}
    \end{subfigure}\medskip\\
    \begin{subfigure}[t]{\linewidth}
      \centering%
      \includegraphics[width=0.6\linewidth,trim={10cm 9cm 0cm 0.24cm},clip]{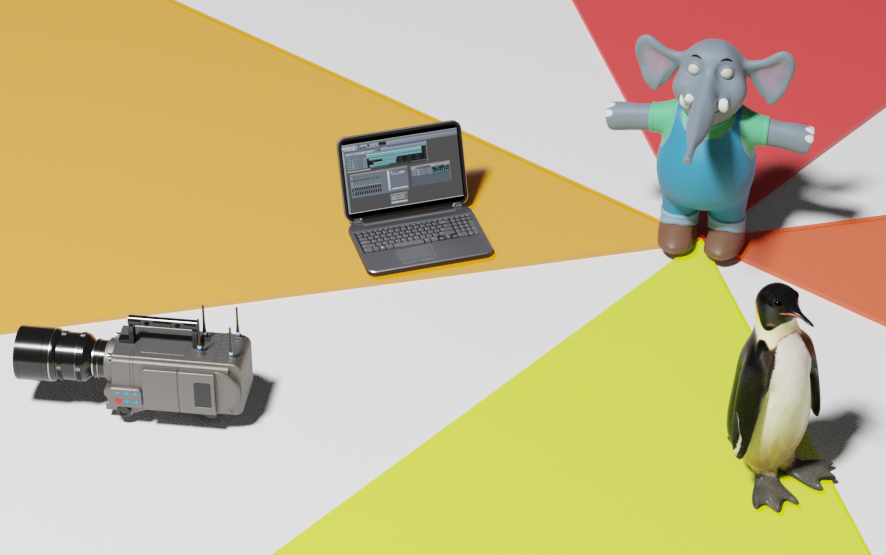}%
      \caption{Relation identification\label{fig:subtask3}}
    \end{subfigure}
  \end{minipage}
  \hfill{}
  \caption{We propose a VQA dataset for grounding relative directions; in addition to questions on relative directions, it encompasses questions regarding object detection and orientation estimation.
  \label{fig:pipeline}}
\end{figure}

\begin{table*}[t]
\small
    \centering
    \begin{tabular}{lccccccc}
    \toprule
        \multirow{2}{*}{Dataset} &
        \multirow{2}{*}{\begin{tabular}{c}\#Samples\end{tabular}} &
        \multirow{2}{*}{\begin{tabular}{c}\#Images\end{tabular}} &
        \multirow{2}{*}{\begin{tabular}{c}Automatic\\generation\end{tabular}} &
        \multirow{2}{*}{\begin{tabular}{c}Non-abstract\\objects\end{tabular}} &
        \multirow{2}{*}{\begin{tabular}{c}Multiple\\tasks\end{tabular}} &
        \multirow{2}{*}{\begin{tabular}{c}Varying object\\counts\end{tabular}} &
        \multirow{2}{*}{\begin{tabular}{c}Relative\\directions\end{tabular}} \\
        ~ & ~ & ~ & ~ & ~ & ~ & ~& ~ \\ \midrule
        CLEVR~\cite{johnson_clevr_2017}           & 1M & 100k & \cmark & \xmark & \cmark & \cmark & \xmark \\
        PTR~\cite{hong_ptr_2021}             & 700k & 70k & \cmark & \cmark & \cmark & \cmark & \xmark \\
        SpatialSense~\cite{yang_spatialsense_2019}    & 17.5k & 11.6k & \xmark & \cmark & \xmark & \cmark & \xmark \\
        Rel3D~\cite{goyal_rel3d_2020}           & 27k & 27k & \xmark & \cmark & \xmark & \xmark & \cmark \\ \midrule
        GRiD-3D \textit{(ours)}         & 445k & 8k & \cmark & \cmark & \cmark & \cmark & \cmark \\ \bottomrule
    \end{tabular}
    \caption{Comparison of GRiD-3D with existing spatial reasoning benchmarks.}
    \label{table:datasetcomparison}
\end{table*}

Talking about the locations of objects is a basic communicative skill for humans to manage their everyday activities \cite{bloom_language_1999,coventry_saying_2004}. Among several ways of talking about locations, humans often use the relative direction of the target object from the perspective of a reference object. For example, to describe the target object \textit{laptop} in Fig. \ref{fig:horse_penguin} one can take the perspective of the reference object \textit{elephant} and say that ``the \textit{laptop} is \textit{to the right} of the \textit{elephant}''. As relative directions are common in everyday use of language, they are important in human-robot interaction and artificial intelligence \cite{moratz_spatial_2006,lee_starvars:_2013,hua_qualitative_2018}.

This paper is concerned with grounding relative directions with end-to-end neural networks. This task is not trivial since, following human intuition, it is composed of a sequence of three subtasks: 
Given a collection of objects in a scene,
it first has to detect the objects by binding the visual 
and the linguistic representation of the target and the reference object in the scene (cf. Fig.~\ref{fig:subtask1}). Second, it has to estimate the orientation of the reference object (cf. Fig.~\ref{fig:subtask2}). Finally, it has to determine the direction of the target object based on the orientation of the reference object (cf. Fig.~\ref{fig:subtask3}).

Conventional approaches to process the previous three subtasks would hand-design 
a pipeline of modules that carry out each subtask. Such pipeline-based modular approaches often perform well, in particular when the target task is well-known such that the modules can be tailored to the subtasks of the target task. They also often facilitate interpretation and diagnosis of the results. However, they cannot be flexibly applied to a new task without either training every module on the new task or adding new modules and re-configuring the pipeline. End-to-end differentiable neural models, on the other hand, do not suffer from the latter deficiency of the modular approaches. Therefore, investigating how well such end-to-end neural models ground relative directions is an important and interesting open research question.

We view the relative direction grounding task from the perspective of \emph{visual question answering} (VQA)~\cite{wu_visual_2017} and propose a dataset-based approach that allows existing end-to-end differentiable neural VQA models to implicitly learn to solve the three subtasks. As we can \emph{control} with question inputs how the neural VQA models process images, we allow the models to solve the three subtasks merely by including additional VQA questions in the dataset pertaining to all three subtasks, \emph{without} any modifications of the model architectures. The contributions of this paper are the following:
\begin{enumerate}
  \item We introduce GRiD-3D\footnote{\url{https://github.com/knowledgetechnologyuhh/grid-3d}}, a novel diagnostic VQA dataset for end-to-end relative direction learning. This dataset is more realistic than existing synthetic datasets for visual reasoning. It includes objects from different real-world categories with intrinsic orientations, which allows us to evaluate the capabilities of existing neural models for grounding relative directions.
  \item We conduct extensive experiments on our dataset and show that the state-of-the-art generic neural VQA models FiLM~\cite{perez_film_2018} and MAC~\cite{hudson_compositional_2018} are capable of grounding relative directions. We demonstrate that the models learn the three subtasks that are intuitively necessary for grounding relative directions, i.e., (i) object detection, (ii) pose estimation, and (iii) relation identification. We also identify that the three subtasks are learned in this specific order, which suggests the presence of an emergent curriculum, i.e., the three subtasks are ordered in increasing difficulty, and easier tasks facilitate learning more difficult tasks.
\end{enumerate}

\section{Related Work}
\definecolor{myBlue}{RGB}{78,118,166}
\begin{figure*}[t]
  \begin{subfigure}[t]{0.25\linewidth}
    \centering%
    \includegraphics[width=0.78\linewidth]{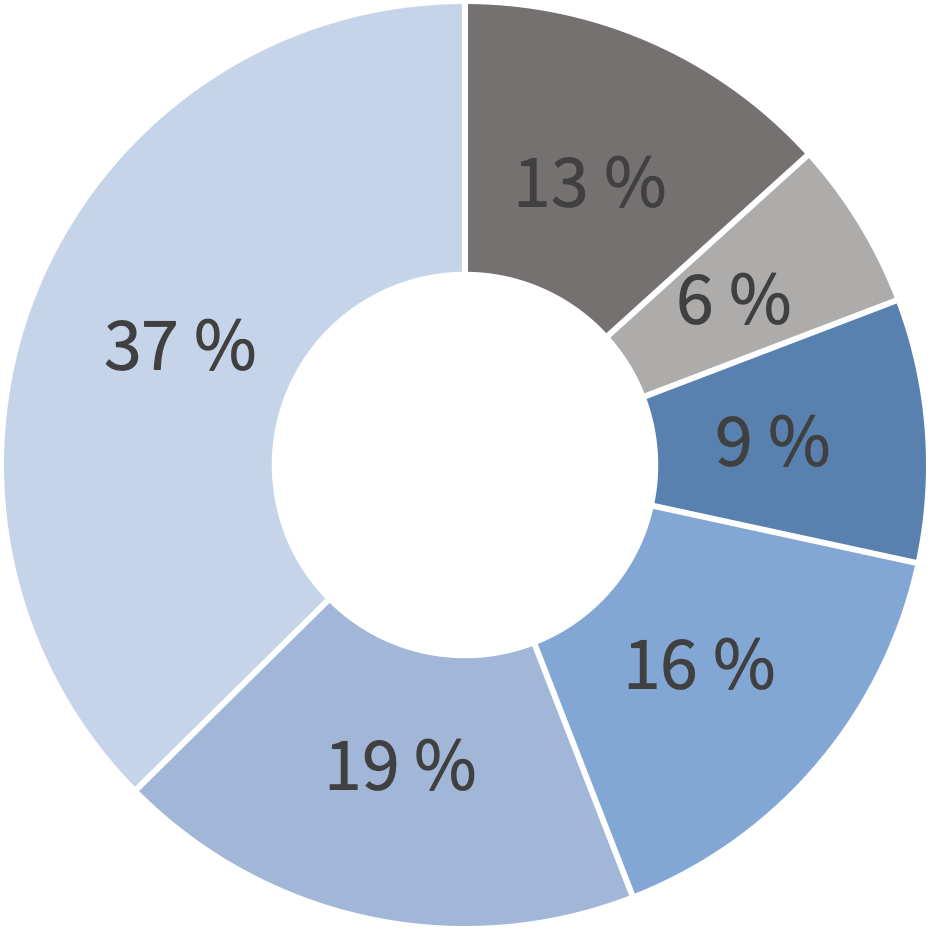}
  \end{subfigure}%
  \hfill
  \begin{subfigure}[t]{0.29\linewidth}
    \centering%
    \includegraphics[width=\linewidth]{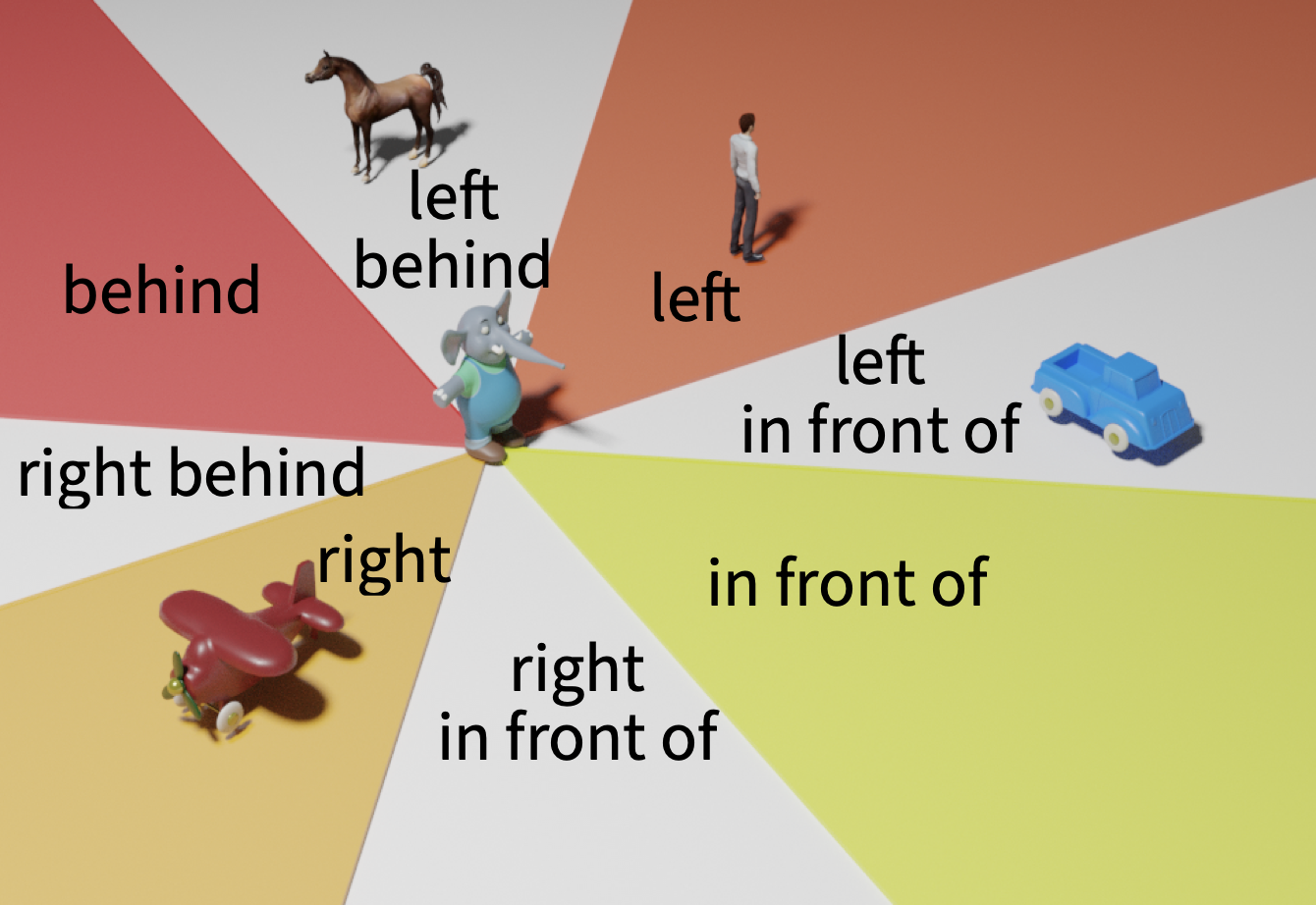}
  \end{subfigure}%
  \hfill
  \begin{subfigure}[t]{0.31\linewidth}
    \centering
    \includegraphics[width=\linewidth]{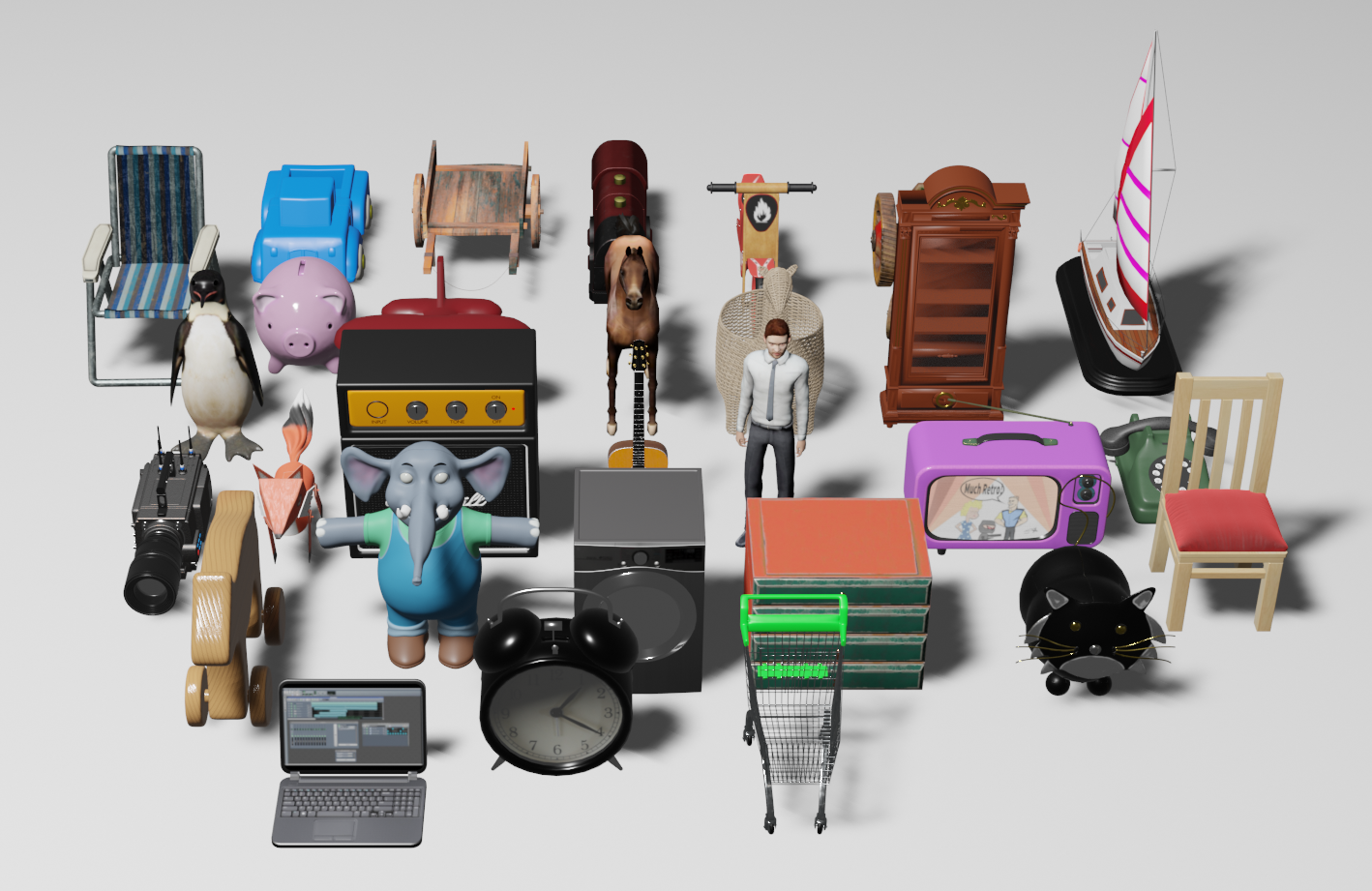}
    \label{fig:allObjects}
  \end{subfigure}%
  \hfill{}
  \medskip\\
  \begin{subfigure}[t]{1.0\linewidth}
    \centering
    \includegraphics[width=\linewidth]{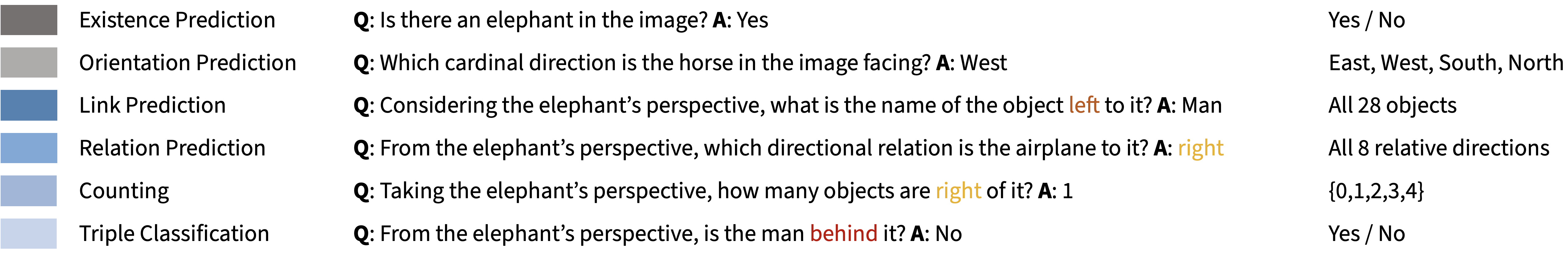}
  \end{subfigure}
  \caption{\textbf{Top left}: Distribution of per-task questions. Tasks shown in \colorbox{myBlue}{\textbf{\color{white}blue}} require \emph{understanding relative directions}. \textbf{Top center}: Overview of the 8 different relative directions from the viewpoint of the reference object \textit{elephant}. \textbf{Top right}: All 28 3D-objects used for the synthetic dataset. Each object features a clear front side. \textbf{Bottom}: From left to right: the six tasks; sample questions and answers for each task; the answer set.}
  \label{fig:taskSummaryOverview}
\end{figure*}

\paragraph{General relation learning}

Grounding relative directions involves situated perception, such as vision, in addition to language. A prominent task in the area of vision and language integration is \emph{visual question answering} (VQA), which is concerned with answering questions about an image~\cite{wu_visual_2017}. Several datasets have been proposed for VQA~(e.g., \cite{goyal_making_2017,johnson_clevr_2017,suhr_corpus_2019,hudson2019gqa}).
The existing VQA datasets, however, typically use spatial relations in an absolute frame of reference (i.e., \emph{left} and \emph{right} correspond to the \emph{left} and \emph{right} side of the input image) and do not offer a way to evaluate a model's capability in dealing with relative directions.


\paragraph{Relative directions}
In the area of knowledge representation and reasoning, much attention has been paid to inference with relative direction information~\cite{moratz_representing_2006,lee_starvars:_2013,hua_qualitative_2018,freksa_formal_2018}. 
Existing work in this area typically assumes that the primitives, i.e., the labels of objects and the relations between objects, are given, and does not address the implicit learning of such information based on sensory stimuli from, e.g., images.


\paragraph{VQA Datasets}
The development and evaluation of powerful VQA models relies 
 on the availability of well-curated datasets. 
Designing such datasets of real-world images is labor-intensive; for a large-scale dataset, this can only be done by using pretrained object detectors, by crowdsourcing the annotation effort 
\cite{krishna_visual_2017}
or by leveraging existing annotations, e.g., in the form of narrated videos \cite{yang2021just}.
An alternative are synthetic datasets generated with 3D rendering environments. While artificially generated images do not provide the full complexity and noisiness of real-world images, they do offer several advantages: (i) images can be generated in an unbiased way, e.g., with regard to the distribution of relevant objects, relations and similar properties; (ii) images can be automatically generated with all required annotations; (iii) such datasets can easily be upscaled based on the demands of the task and the neural architecture. They are therefore well-suited as diagnostic datasets.

\paragraph{Spatial relation learning with VQA datasets}
Recently, several datasets have been proposed that overcome the limitations of existing VQA datasets in dealing with spatial relations. The PTR dataset~\cite{hong_ptr_2021} focuses on the part-whole relationships between entities in synthetic scenes. 
However, the dataset does not deal with relative directions, i.e., the directions from the perspective of the reference objects.

SpatialSense~\cite{yang_spatialsense_2019} is a crowdsourced dataset, where human annotators provided spatial relation labels that are difficult to predict using simple cues (e.g., 2D spatial configurations or language priors). Annotators made extensive use of relative directions, which turned out to be major failing cases for the baseline models.

Rel3D~\cite{goyal_rel3d_2020} is also a crowdsourced dataset, but different from SpatialSense, the images in Rel3D are synthetically generated with 3D ground-truth information. Rel3D provides examples with reduced annotation bias. 
The dataset is limited in that the images in the dataset contain only two objects, and it provides only one type of question, i.e., predicting the correctness of (object\textsubscript{1}, relation, object\textsubscript{2}) triples. By contrast, our dataset provides a varying number of objects in each scene and different categories of questions. 

Table \ref{table:datasetcomparison} shows an overview of relevant VQA datasets on relation learning. While all datasets in the table include spatial relations in their questions, GRiD-3D is the only dataset that features questions on relative directions as part of a more extensive, multi-task question inventory. With GRiD-3D, we address this gap in the state of the art of existing VQA datasets.

\section{GRiD-3D VQA Dataset}
We introduce GRiD-3D (\underline{\textbf{G}}rounding \underline{\textbf{R}}elat\underline{\textbf{i}}ve \underline{\textbf{D}}irections in \underline{\textbf{3D}}), a novel diagnostic VQA dataset that enables learning relative directional relations between objects. Our dataset comprises \mbox{8\,000} synthetic images and \mbox{445\,080} questions addressing six different reasoning tasks: \task{Existence Prediction}, \task{Orientation Prediction}, \task{Link Prediction}, \task{Relation Prediction}, \task{Counting}, and \task{Triple Classification}. Exemplary input questions and answers per reasoning task are shown in Fig.~\ref{fig:taskSummaryOverview}.

The images are split in an 80:10:10 ratio without overlapping images between training, validation and test sets (i.e., \mbox{6\,400}, 800, and 800, respectively). The \mbox{445\,080} questions are split into similar proportions (i.e., \mbox{357\,839}, \mbox{45\,030}, and \mbox{42\,211}, respectively).

All images have a 480x320 pixel resolution and are rendered using Blender\footnote{\url{https://www.blender.org/}} by randomly placing 2--5 objects onto a plane in a non-overlapping manner and following a uniform distribution. We choose a consistent lighting setup that provides shadows for the sake of more realistic scenes. In addition, we restrict the image generation to a fixed camera angle that allows for partial but not complete occlusion of objects. Consequently, we have one image per scene, and we will therefore use these two terms interchangeably in this work. 

Scene objects are randomly selected from a set of 28 different non-abstract 3D models that all have a distinctive front side, thus allowing to describe a spatial layout with respect to the object’s intrinsic frame of reference, i.e., via relative directions. All 3D models of the objects were obtained from BlenderKit\footnote{\url{https://www.blenderkit.com/}}, released under “Royalty-Free” or “CC0” licenses. They are depicted in Fig.~\ref{fig:taskSummaryOverview}.

The dataset obtains ground-truth annotations regarding absolute position, orientation, and relative directions of objects for each generated scene. Following the strategy of~\citeauthor{johnson_clevr_2017}~[\citeyear{johnson_clevr_2017}], questions are expressed as a functional program on each scene's ground-truth information. Each task's questions, instantiated and validated based on depth-first search, follow a uniform answer distribution. For example, the body of questions addressing the object \task{Existence Prediction} task has a 50/50 ratio of \textit{yes} or \textit{no} as the target answer. For each reasoning task, we vary question phrasing by using different question skeletons as well as by omitting or replacing terms by corresponding synonyms with some predefined probability.

The ratio between per-task question counts, as illustrated in Fig.~\ref{fig:taskSummaryOverview}, 
is drawn from the ratio of the total number of possible questions per task. For example, the number of uniquely instantiable questions on \task{Orientation Prediction} per scene equals the total number of its objects, whereas, for the same scene, we can generate twice as many questions on \task{Existence Prediction} (one negative and one positive sample per object).

\section{Language-Controlled Multi-Task Learning}

\begin{figure}[t]
  \centering%
  \hfill
  \begin{subfigure}[t]{0.47\linewidth}
    \centering%
    \includegraphics[width=\linewidth]{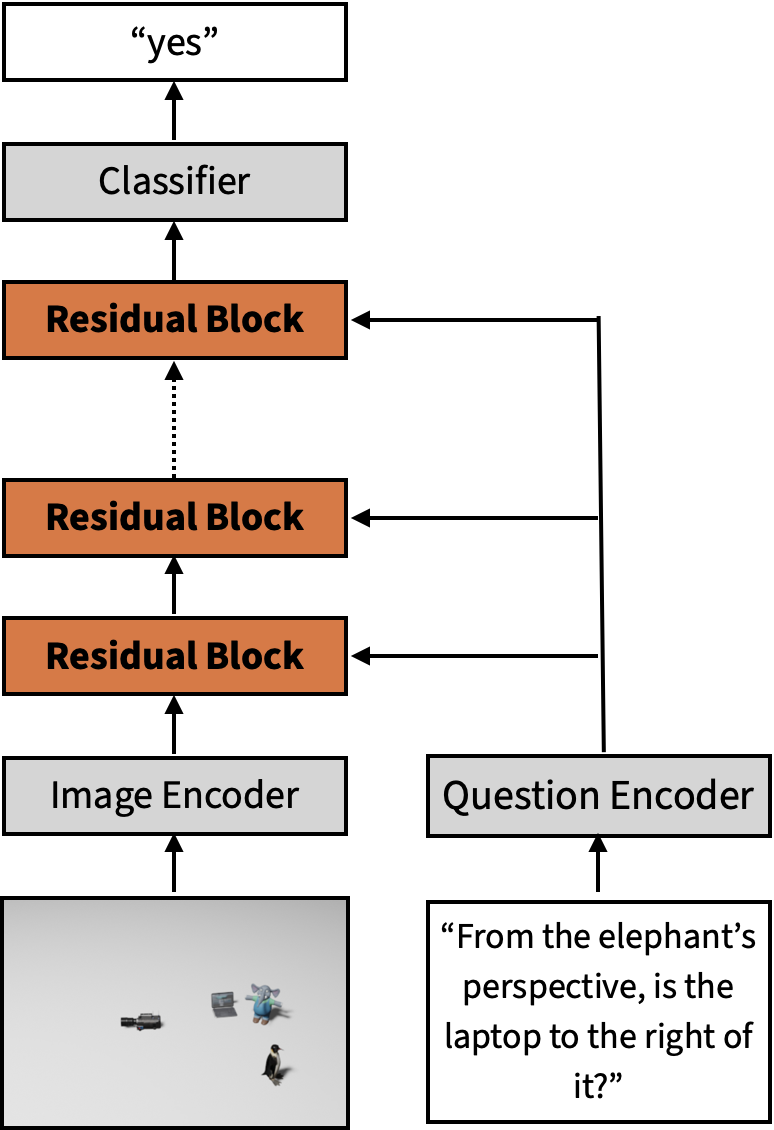}
    \caption{FiLM}
  \end{subfigure}
  \hfill
  \begin{subfigure}[t]{0.47\linewidth}
    \centering%
    \includegraphics[width=\linewidth]{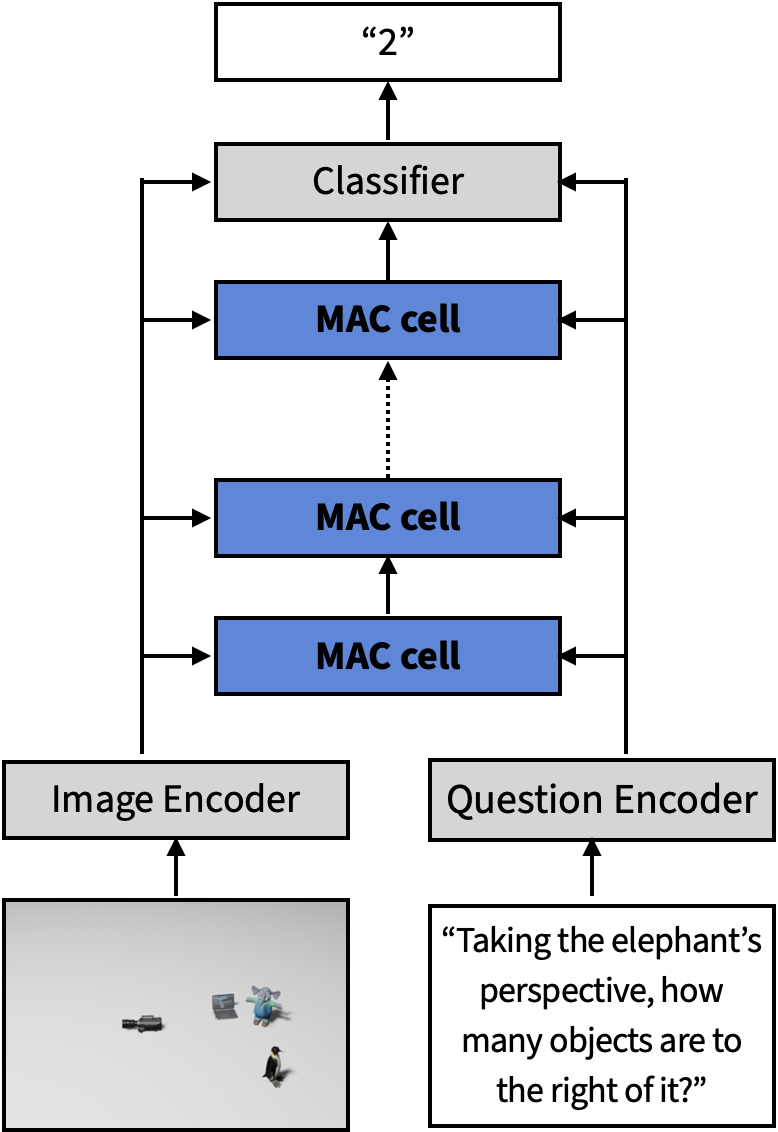}
    \caption{MAC}
  \end{subfigure}
  \hfill
  \caption{An abstract overview of the FiLM and MAC architectures. In both cases, the encoded question is passed to each generic unit (colored in orange and blue, respectively) and affects how the encoded image is processed by the models.\label{fig:filmandmac}}
\end{figure}

Solving VQA requires different capabilities of a model, as the questions in a VQA dataset involve multiple tasks, e.g., object detection, matching, comparison, counting, relation identification. For example, to answer the following question in the CLEVR dataset~\cite{johnson_clevr_2017}: ``There is a sphere with the same size as the metal cube; is it made of the same material as the small red sphere?'' a model needs to be able to perform object detection, matching, and comparison. 

Modular VQA models, such as the ones proposed by \citeauthor{johnson_inferring_2017}~[\citeyear{johnson_inferring_2017}], \citeauthor{hu_learning_2017}~[\citeyear{hu_learning_2017}], \citeauthor{yi_neural-symbolic_2018}~[\citeyear{yi_neural-symbolic_2018}], achieve multi-task learning by providing a module for each subtask comprising a target task and automatically building pipelines based on those modules. The modules are hand-crafted (sometimes with trainable parameters) and require detailed knowledge about the target task. It is, therefore, difficult, if not impossible, to apply them to new tasks without designing new such modules.

By contrast, end-to-end differentiable VQA models, such as FiLM~\cite{perez_film_2018} and MAC~\cite{hudson_compositional_2018}, do not need specific modules designed for specific subtasks. Instead, those models feature generic units called \emph{residual blocks} (FiLM) or \emph{MAC cells} (MAC), which are controlled by the language input features.

As can be seen in the high-level overview of the model architectures in Fig.~\ref{fig:filmandmac}, each generic unit receives as input the output from the previous unit together with the encoded question (and, in the case of MAC, the encoded image). Therefore, the encoded question basically controls each generic unit and modifies their inner workings such that the whole architecture can appropriately process the image input and accomplish the target task.\footnote{There are two main differences between FiLM and MAC: (i) in FiLM the encoded image is passed to the first residual block only, whereas in MAC the encoded image is passed to every MAC cell; (ii) the residual blocks do not share weights, while the MAC cells share weights; thus MAC can be regarded as a recurrent model with a fixed depth of recurrence.}



\section{Evaluations}
\label{sec:evaluations}

In this section, we evaluate the two state-of-the-art end-to-end differentiable VQA models, FiLM and MAC. 
As both models have their generic units directly controlled by the input question, each unit's function can be adjusted with respect to the target task and can process the input image accordingly. 

Following our question of whether a model can ground relative directions, we pay particular attention to model performance on those tasks targeting such capability, i.e. \task{Link Prediction}, \task{Relation Prediction}, \task{Counting}, and \task{Triple Classification}.
We choose the accuracy on \task{Triple Classification}, which comprises the largest part of the dataset (cf.~Fig.~\ref{fig:taskSummaryOverview}), as a representative evaluation criterion for grounding relative directions and answer the following questions:
\begin{compactenum}
  \item Is multi-task learning necessary for \task{Triple Classification}?
  \item Is multi-task learning with the tasks in GRiD-3D sufficient for \task{Triple Classification}?
  \item Is there an order in learning the tasks?
  \item What tasks in GRiD-3D are necessary for \task{Triple Classification}?
\end{compactenum}
Our evaluation framework is implemented using the PyTorch\footnote{\url{https://pytorch.org/}} library and existing implementations of FiLM and MAC with their default parameter configurations used for the evaluations on the CLEVR dataset, except for the number of MAC cells that is set to four instead of the default value 12, as we observed an improved model performance. We run each experiment three times for 50 epochs. Each graph of an experiment shows the \emph{mean} and the \emph{standard deviation} of the three runs.

\subsection{Multi-Task Learning is Necessary}

\begin{figure}[t]
  \centering
  \begin{subfigure}[t]{0.48\linewidth}
    \centering
    \includegraphics[width=\linewidth]{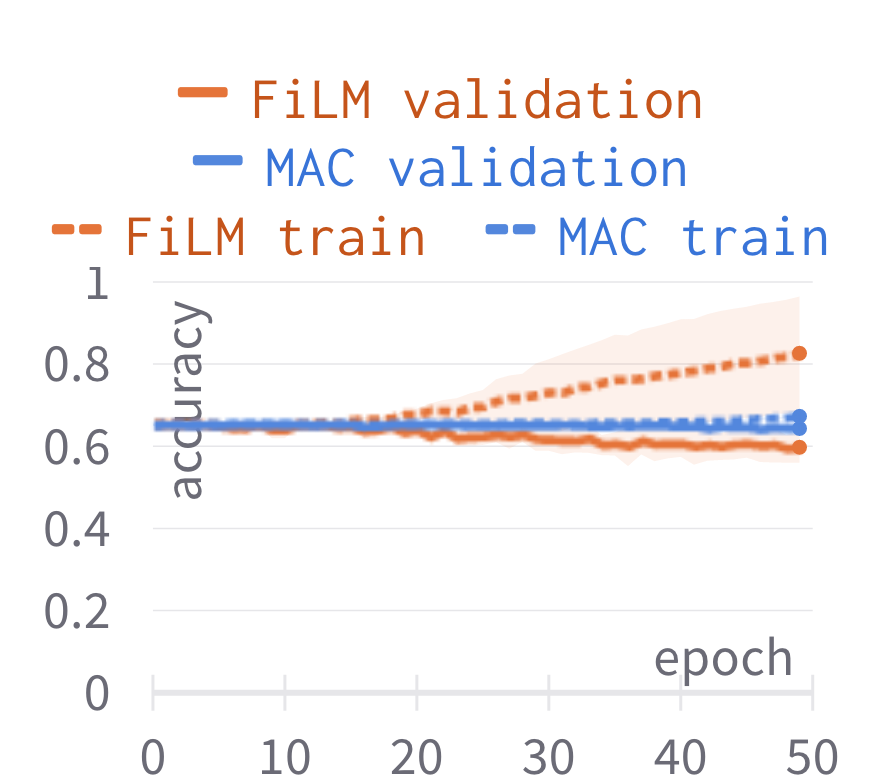}
    \caption{Trained on \task{Triple Classification} only.\label{fig:single-task}}
  \end{subfigure}
  \hfill
  \begin{subfigure}[t]{0.48\linewidth}
    \centering
    \includegraphics[width=\linewidth]{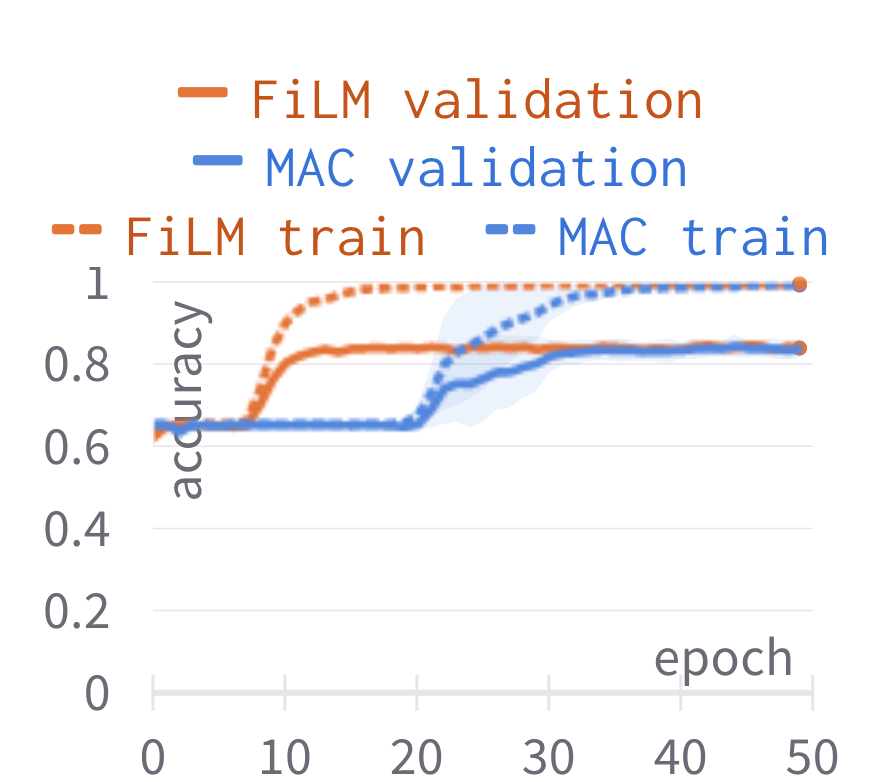}
    \caption{Trained on all tasks in GRiD-3D.\label{fig:multi-task}}
  \end{subfigure}
  \caption{Mean validation and training accuracies of FiLM and MAC on \task{Triple Classification}.}
\end{figure}

\begin{figure*}[t]
  \begin{subfigure}[t]{0.48\textwidth}
    \includegraphics[width=0.495\linewidth]{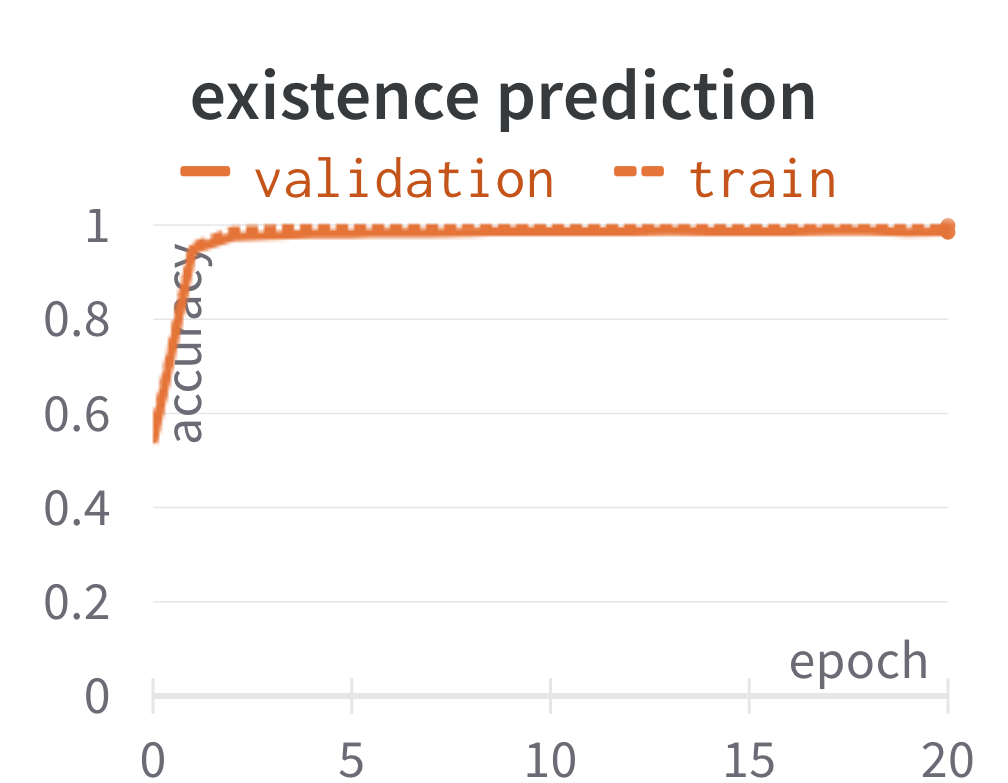}%
    \includegraphics[width=0.495\linewidth]{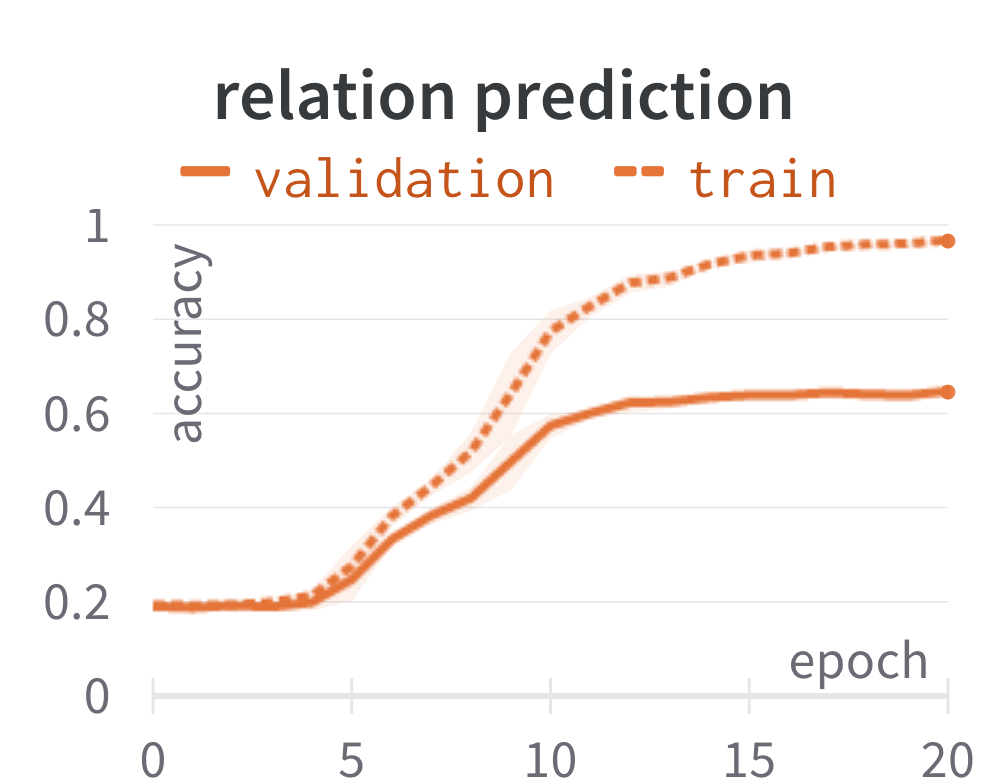}%
    \\
    \includegraphics[width=0.495\linewidth]{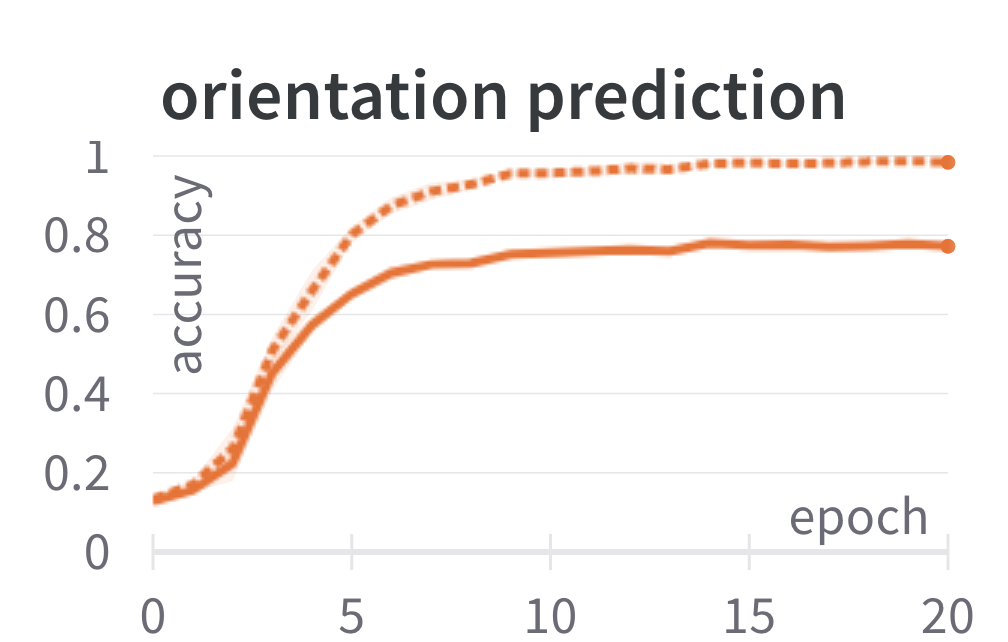}%
    \includegraphics[width=0.495\linewidth]{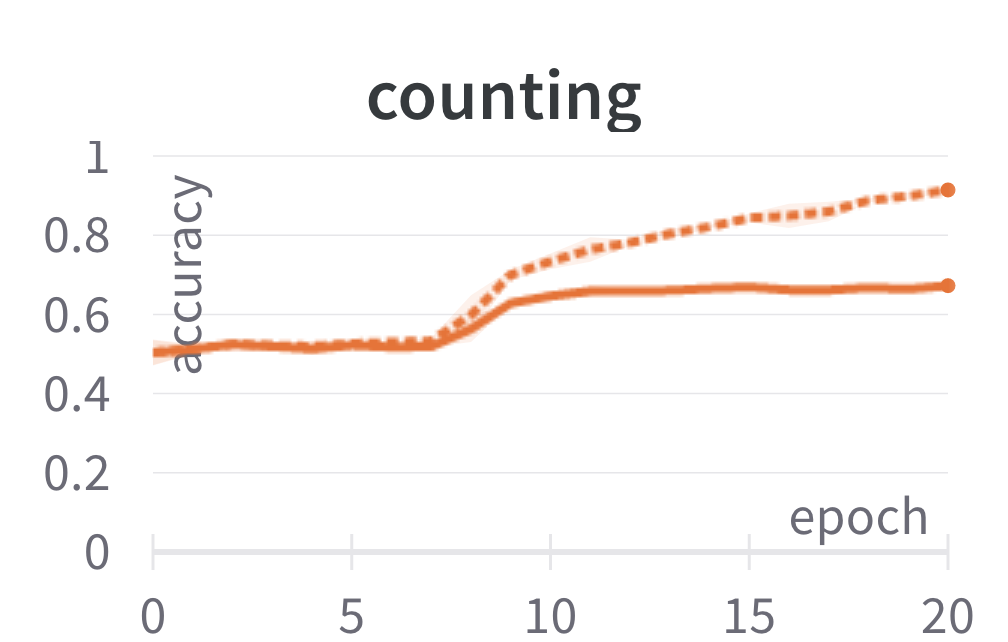}%
    \\
    \includegraphics[width=0.495\linewidth]{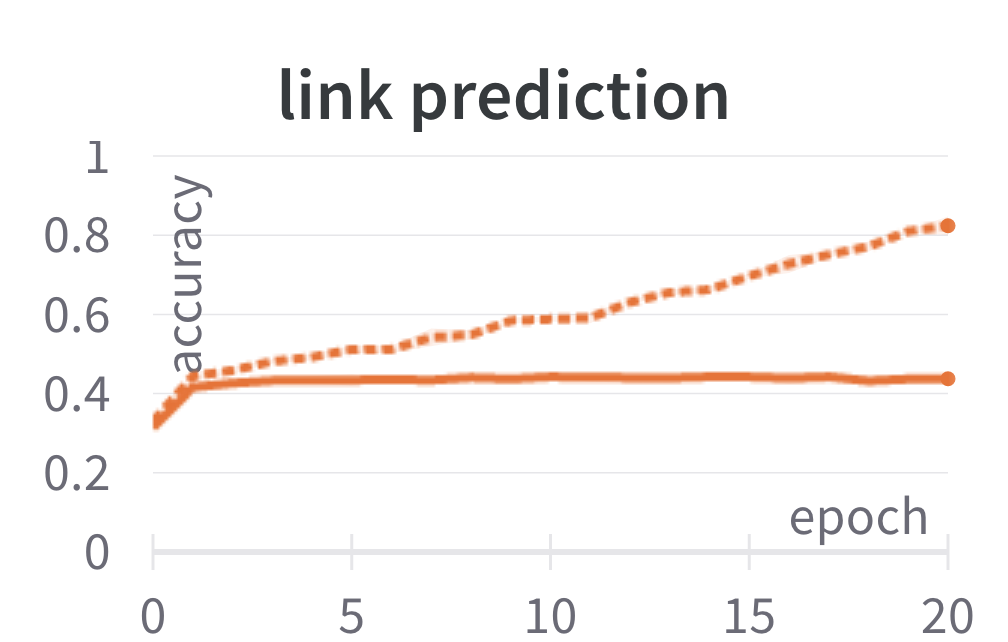}%
    \includegraphics[width=0.495\linewidth]{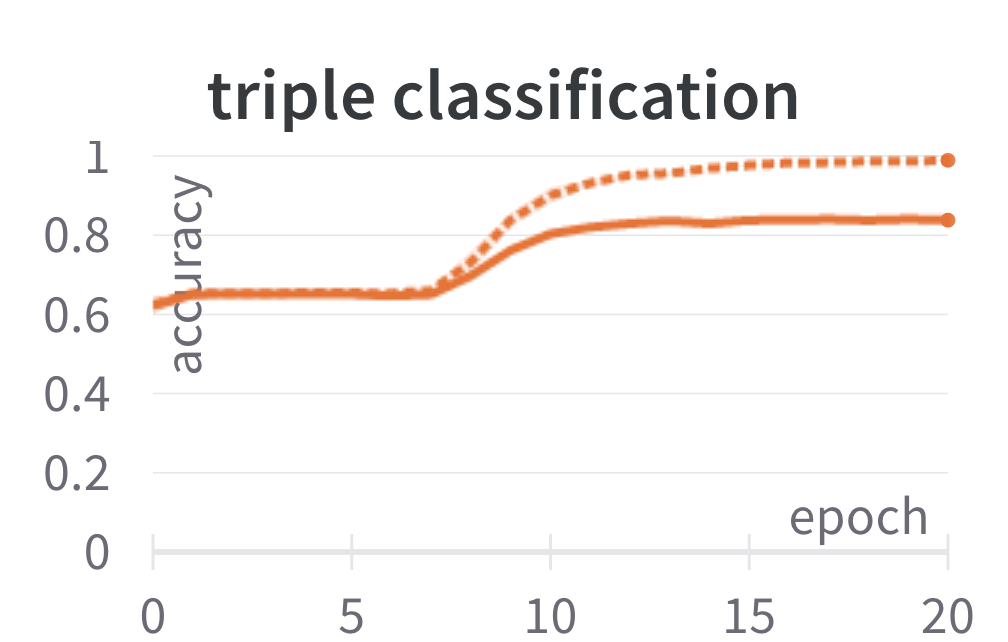}%
    \caption{FiLM}
  \end{subfigure}
  \hfill
  \begin{subfigure}[t]{0.48\textwidth}
    \includegraphics[width=0.495\linewidth]{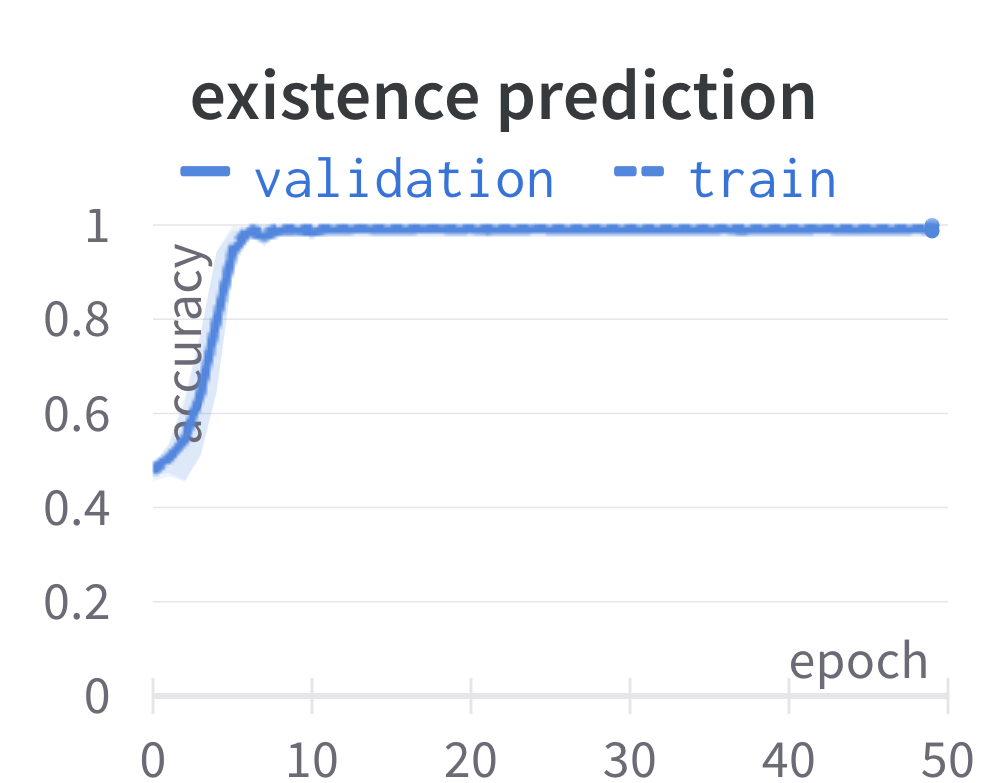}%
    \includegraphics[width=0.495\linewidth]{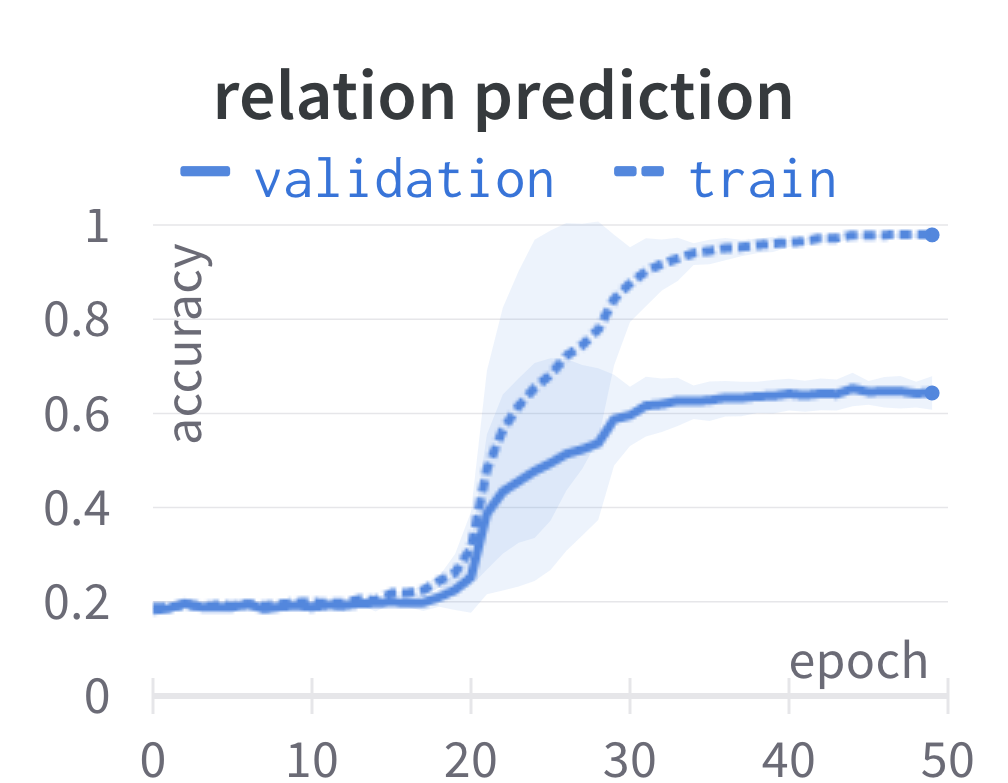}%
    \\
    \includegraphics[width=0.495\linewidth]{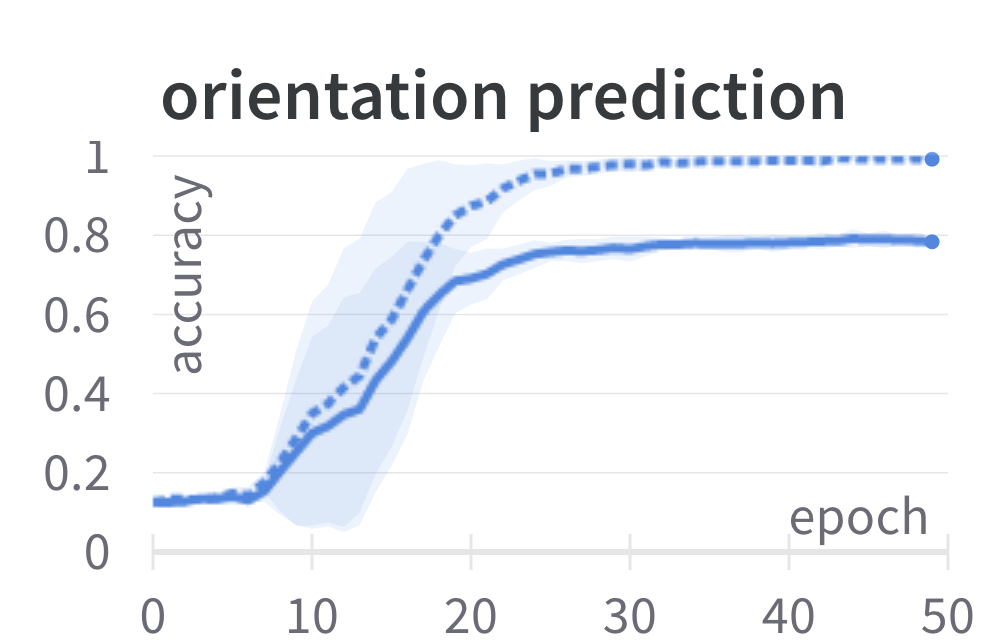}%
    \includegraphics[width=0.495\linewidth]{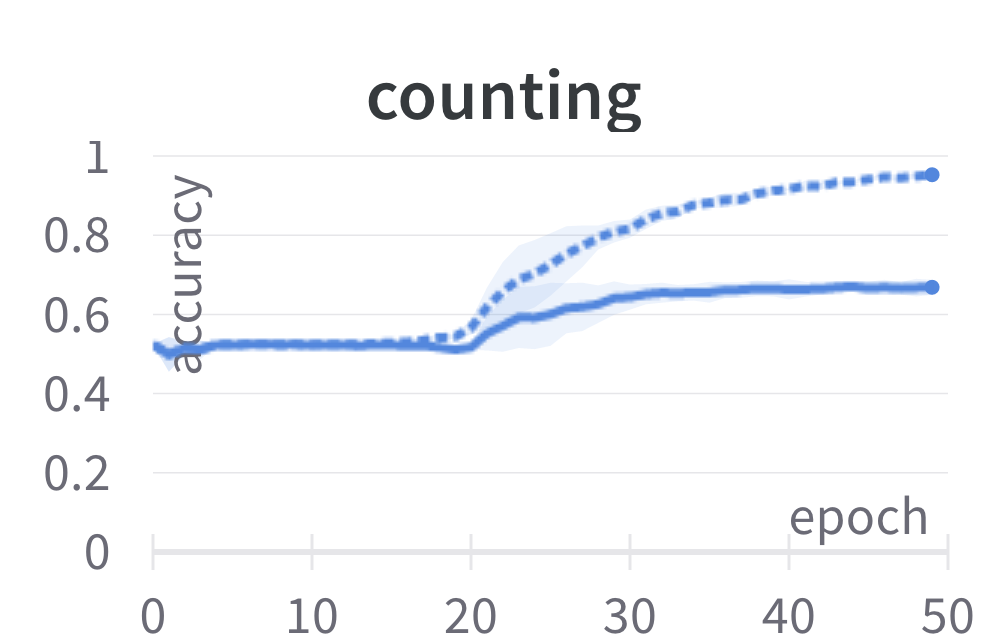}%
    \\
    \includegraphics[width=0.495\linewidth]{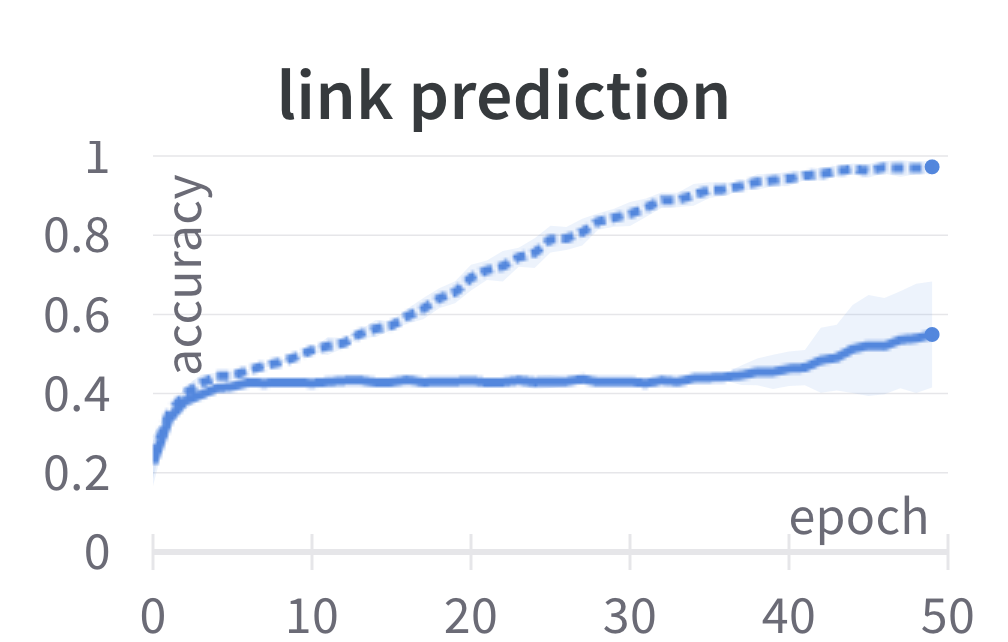}%
    \includegraphics[width=0.495\linewidth]{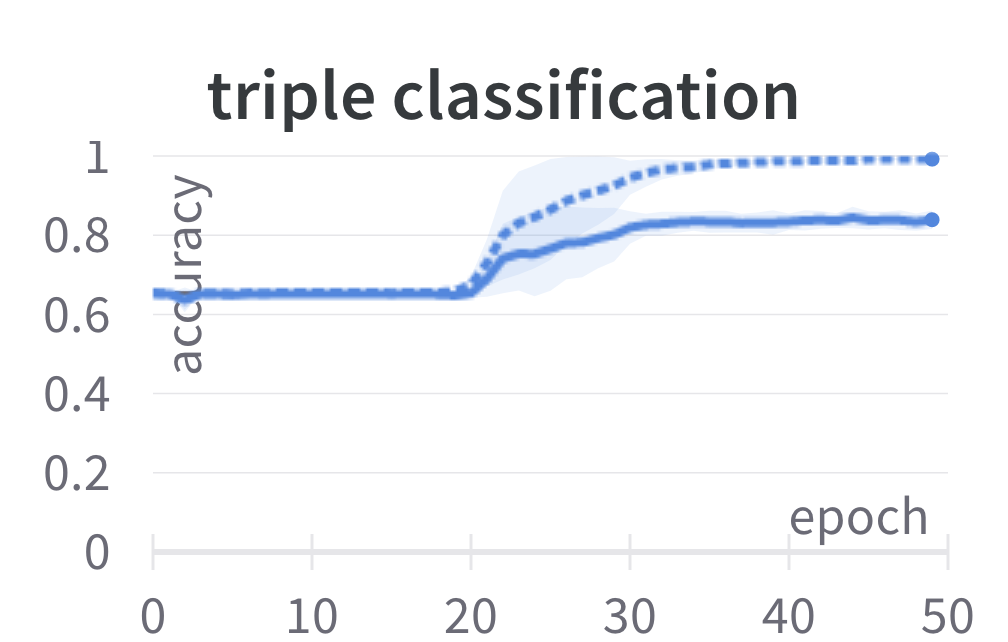}%
    \caption{MAC}
  \end{subfigure}
  \caption{Multi-task learning results on all six tasks of the GRiD-3D dataset. For FiLM, only the first 20 of the 50 total training epochs are shown, to allow for comparison of the two models in terms of convergence behavior. \label{fig:learning-dynamics}}
\end{figure*}

The first intuitive question we can ask is whether multi-task learning is necessary at all to solve the \task{Triple Classification} task from GRiD-3D. To answer this question, we train and evaluate FiLM and MAC on \task{Triple Classification} only. The training and validation learning curves of both models of this experiment are shown in Fig.~\ref{fig:single-task}. While the initial mean validation accuracies of the two models are around 65\% after the first epoch, they drop to 60\% and 64\% after 50 epochs of training. This is a rather poor performance considering that a random baseline achieves 50\% accuracy.

We can make two further observations: First, the mean validation accuracies are about 10\% above the chance level, indicating some spurious correlations in the dataset that the models can exploit. Second, both models even failed to consistently learn the training data for \task{Triple Classification}, as the mean training accuracies after 50 epochs are 83\% and 67\%. This hints at the fact that \task{Triple Classification} is a complex task, and merely memorizing some correlations between the inputs and the outputs is not sufficient. Overall, the result shows that single-task learning is not sufficient and hence multi-task learning is necessary for \task{Triple Classification} with GRiD-3D.

\subsection{GRiD-3D Tasks are Sufficient}

Based on our previous finding that multi-task learning on GRiD-3D could facilitate \task{Triple Classification},
a natural follow-up question would be whether the tasks included in the GRiD-3D dataset are sufficient to obtain a meaningful estimate of the \task{Triple Classification} capabilities of a neural VQA model.

\begin{wrapfigure}[9]{r}{4cm}
  \hfill
  \includegraphics[width=0.205\linewidth]{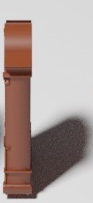} \includegraphics[width=0.205\linewidth]{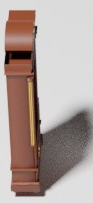}
  \hfill
  \includegraphics[width=0.51\linewidth]{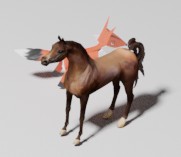}
  \hfill{}
  \caption{Challenging cases: the left and the right side of a closet, and a horse occluding a fox.}
  \label{fig:challenges}
\end{wrapfigure}
To answer this question, we train FiLM and MAC on all six tasks and evaluate their performances on the \task{Triple Classification} task. The results are shown in Fig.~\ref{fig:multi-task}.
In this experiment, both FiLM and MAC achieve 84\% mean validation accuracy after 50 epochs, respectively, improving their mean validation accuracies on the single task experiment by more than 31\%.


Given that there are difficulties in detecting objects (e.g., due to occlusions), estimating orientations (e.g., due to almost symmetric objects such as \textit{closet}) and identifying relations (e.g., target objects being close to the decision boundary between two relations), further improving the performance would be challenging without injecting additional prior knowledge to the models (cf.~Fig.~\ref{fig:challenges}). Consequently, we can conclude that the tasks in GRiD-3D are sufficient to help FiLM and MAC solve the \task{Triple Classification} task.

\subsection{The Order of Learning the Tasks}

Multi-task learning with the tasks in GRiD-3D allows FiLM and MAC to solve \task{Triple Classification} subject to the intrinsic difficulties of the task mentioned previously. It is, however, not clear \emph{why} and \emph{how} adding different question-answer pairs (e.g., questions about orientations), while keeping the same set of images in the training set, leads to the sudden improvement both in the training and the validation performances. We can partially answer this question by observing the validation performances on all six  tasks that are
presented in Fig.~\ref{fig:learning-dynamics}. We observe that there is an order of the tasks that governs the dynamics of learning, which reflects the pipeline of the subtasks for grounding relative directions as suggested in the introduction of the paper (cf.~Fig.~\ref{fig:pipeline}):
\begin{compactenum}
  \item \task{Existence Prediction} is learned, which corresponds to learning object detection (cf.~Fig.~\ref{fig:subtask1})
  \item \task{Orientation Prediction} is learned, which corresponds to learning orientation estimation (cf.~Fig.~\ref{fig:subtask2}),
  \item The remaining four tasks that involve grounding relative directions are learned (with the exception of link prediction in the case of FiLM), which correspond to learning relation identification (cf.~Fig.~\ref{fig:subtask3}).
\end{compactenum}
This intuitive but surprising emergence of an implicit order of tasks suggests that curating a VQA dataset with well-chosen questions can already facilitate existing end-to-end differentiable VQA models such as FiLM and MAC to solve new tasks that they are not initially designed for and are difficult to solve in isolation, all \emph{without} including any additional images.

One noticeable result in Fig.~\ref{fig:learning-dynamics} is the discrepancy between the mean validation and training performance on \task{Link Prediction}. After 50 epochs, FiLM and MAC achieve 44\% and 55\% mean validation accuracies despite their mean training accuracies being as high as 97\% (not shown in the figure for FiLM). Furthermore, the mean validation accuracies of both models quickly reach 42\%, which is very high considering the fact that there are 28 candidate target objects the models can choose from. Because their performances on \task{Existence Prediction} peak about the same time, it is likely that the models have learned to first \emph{detect} the reference object and reduce the number of candidate objects from 27 to the remaining 2.5 ones in each scene (there are on average 3.5 objects in each scene), leading to 40\% random chance to guess the target object correctly. These observations show that (i) \task{Link Prediction} is intrinsically more difficult than other tasks, (ii) MAC is more robust in learning \task{Link Prediction}, and (iii) not all tasks in GRiD-3D are necessary for solving \task{Triple Classification}.

\subsection{Essential Tasks for Grounding Relative Directions}

\begin{figure}[t]
  \centering
  \hfill
  \begin{subfigure}[t]{0.48\linewidth}
    \centering \includegraphics[width=0.8\linewidth, trim={3.2cm 0.5cm 2.3cm 2.5cm}, clip]{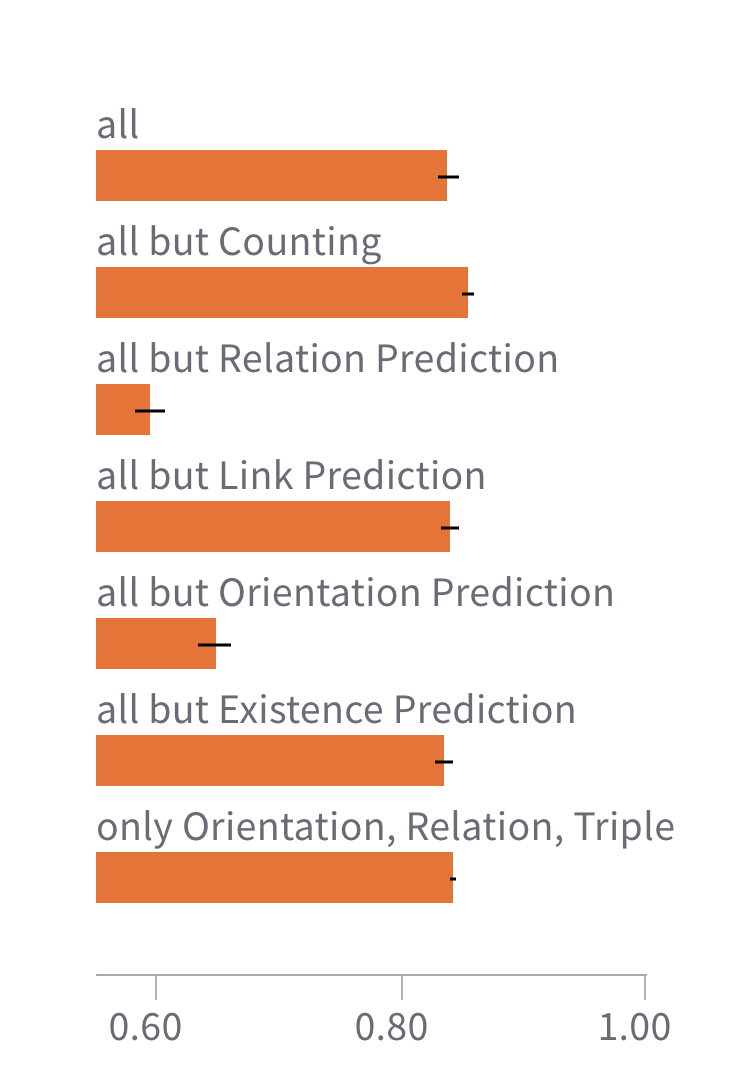}
    \caption{FiLM}
  \end{subfigure}
  \hfill
  \begin{subfigure}[t]{0.48\linewidth}
    \centering \includegraphics[width=0.8\linewidth, trim={3.2cm 0.5cm 2.3cm 2.5cm}, clip]{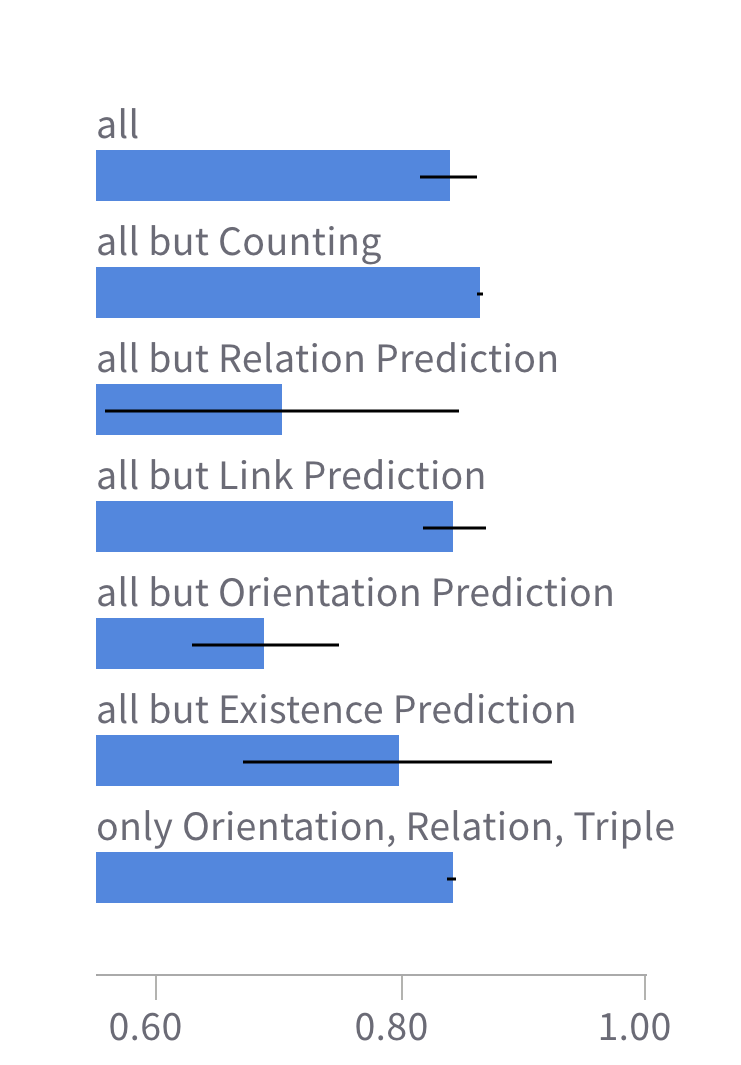}
    \caption{MAC}
  \end{subfigure}
  \caption{Mean validation accuracies of FiLM and MAC on \task{Triple Classification} when trained on selected tasks (see labels).}
  \label{fig:essential-tasks}
\end{figure}

In the preceding subsection, we identified that training FiLM and MAC on all tasks from GRiD-3D is \emph{not} necessary to solve \task{Triple Classification}. This raises the question about what tasks 
are essential. 
To answer this question, we train FiLM and MAC on all but one GRiD-3D tasks to determine the relevance of the one task to \task{Triple Classification}. The results are presented in Fig.~\ref{fig:essential-tasks}.

In the figure, we observe that removing \task{Counting}, \task{Link Prediction}, or \task{Existence Prediction} from the six tasks for training does not lead to much performance drop, whereas removing \task{Orientation Prediction} or \task{Relation Prediction} from the tasks reduces the mean validation accuracy dramatically. On the other hand, training only on the latter two tasks in addition to \task{Triple Classification} already facilitates both models to perform well on \task{Triple Classification}, both achieving 84\% mean accuracies. These findings suggest that \task{Orientation Prediction} and \task{Relation Prediction} are essential for \task{Triple Classification} while the other three tasks are less relevant.

\section{Conclusions}

Grounding relative directions is a relevant and challenging task for visual question answering (VQA). In contrast to absolute directions that require recognizing and localizing objects and determining their relations, relative directions additionally require reasoning about the objects’ orientations. Existing VQA datasets either only address absolute directions or they do not include the subtasks that facilitate grounding relative direction in a multi-task learning setting. 
We address this gap by introducing the novel synthetic VQA dataset \mbox{GRiD-3D}, which focuses on relative directions and features related subtasks of object detection and orientation estimation.

We positively evaluated GRiD-3D with two established neural end-to-end VQA models, FiLM and MAC. We show that these models can learn to correctly answer questions on relative directions when trained with a set of questions pertaining not only to relative directions but also to the subtasks, i.e., object detection and orientation estimation. 
Furthermore, an analysis of the learning process of both models shows how first the subtasks object detection and orientation estimation are learned before questions on relative directions are answered correctly. This incremental learning process based on starting with partial or simplified tasks can be compared to learning during child development and linked to research from cognitive science and deep reinforcement learning (cf.~\cite{kerzel2018accelerating}).
These results support the hypothesis that multi-task learning with an implicit curriculum of subtasks can be beneficial for tackling more difficult VQA tasks. It would be interesting to see the applicability of this learning approach to other tasks that are conventionally tackled with approaches that feature dedicated and hand-designed modules.

GRiD-3D can serve as a diagnostic dataset for the further development of VQA models that 
tackle challenging VQA problems and 
for closing the performance gap for tasks on relative directions in comparison to less complex question classes. 
GRiD-3D will allow investigating further the mechanisms of how multiple tasks support each others' learning, or whether alternative mechanisms can learn relative directions 
in isolation, independent of other tasks. In future work, we will extend the dataset by adding more tasks to allow an even more fine-grained analysis of the curricular learning of models.

\section*{Acknowledgments}
The authors gratefully acknowledge support from the German Research Foundation DFG for the projects CML TRR169, LeCAREbot and IDEAS.


\bibliographystyle{named}
\bibliography{jl,mk}%
\end{document}